# *If Only* We Had Better Counterfactual Explanations: Five Key Deficits to Rectify in the Evaluation of Counterfactual XAI Techniques

Mark T. Keane,[1,2,3] Eoin M. Kenny,[1,3] Eoin Delaney,[1,3] & Barry Smyth[1,2]

[1]School of Computer Science, University College Dublin, Dublin, Ireland
[2]Insight Centre for Data Analytics, Dublin, Ireland
[3]VistaMilk SFI Research Centre, Ireland
{mark.keane, barry.smyth}@ucd.ie, {eoin.kenny, eoin.delaney4}@ucdconnect

## Abstract

In recent years, there has been an explosion of AI research on counterfactual explanations as a solution to the problem of eXplainable AI (XAI). These explanations seem to offer technical, psychological and legal benefits over other explanation techniques. We survey 100 distinct counterfactual explanation methods reported in the literature. This survey addresses the extent to which these methods have been adequately evaluated, both psychologically and computationally, and quantifies the shortfalls occurring. For instance, only 21% of these methods have been user tested. Five key deficits in the evaluation of these methods are detailed and a roadmap, with standardized benchmark evaluations, is proposed to resolve the issues arising; issues, that currently effectively block scientific progress in this field.

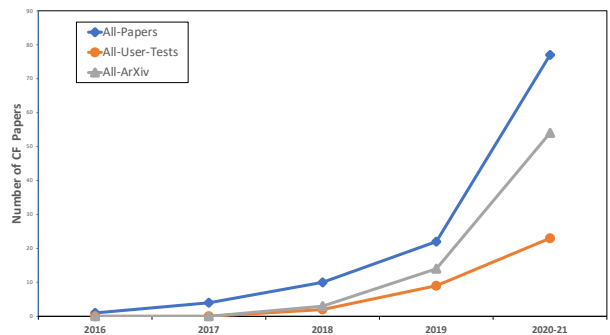

Figure 1: The number of surveyed papers (per annum) on counterfactual XAI (2016-2021) on (a) CF methods (*All-Papers- blue*) (b) CF XAI user studies (*All-User-Tests - orange*) (c) ArXiv from a search of abstracts using the terms "counterfactual explanation" (*All-ArXiv- grey*)[1] [n.b., only a fraction of these papers can be cited in the paper]

## 1 Introduction

In recent years, as AI systems are increasingly deployed in everyday life, there has been a slew of research papers on the problem of eXplainable AI (XAI), driven by concerns about whether these systems are fair, accountable and trustworthy. In the literature on post-hoc explanations, that aim to justify an AI model's predictions *after the fact*, the usefulness of giving counterfactual explanations has gained considerable traction based on claimed technical, psychological and legal benefits (see Figure 1). Counterfactual explanations provide information to users on what might be done to *change* the outcome of an automated decision (e.g., "if your paper had more novelty, it would have been accepted to this conference"). In this paper, we critically review the evaluations carried out on counterfactual explanation methods, focusing on psychological issues. To put it simply, we assess whether there is any evidence that counterfactuals explanations "work" and/or whether the properties promulgated in current methods are relevant to end users. So, this review is, in part, a critique on the paucity of user testing in the area, but it also tries to go beyond this critique to formulate a firmer basis for better evaluation of counterfactual methods in the future (henceforth, we abbreviate the term "counterfactual" as CF).

Obviously, this analysis is also relevant to the bigger picture of XAI, where many have pointed to its failure to properly address user requirements. As in other areas of XAI, CF XAI also suffers from an "over-reliance on intuition-based approaches" that, arguably, impedes scientific progress [Leavitt and Morcos, 2020]. Our main concern is that the XAI community is busily developing technical solutions that may be have no practical benefits to people in real-life. As Barocas et al. [2020] have pointed out, on CF explanations, there is a disconnect between the features of a model and actions in the real-world, that needs to be bridged for these explanation techniques to work successfully.

To be positive, we do not just diagnose the problem but advance solutions in the form of a roadmap for future evaluations, one that argues for standardized, benchmark metrics that are psychologically grounded. Our hope is that these proposals for CF explanation, will also provide a more general template for how XAI might approach user-requirement issues and advance scientific progress.

---

[1] Three seminal CF papers are not shown in this graph: [Nugent *et al.*, 2009; Martens and Provost, 2014; Lim et al, 2009].

In the remainder of this introduction, we briefly consider why counterfactuals have attracted so much attention in XAI, before summarizing the contributions of the current paper. Then, we review the key insights underlie current CF XAI research (see section 2). In the remainder of the paper, we outline five key deficits in CF research that need to be remedied if progress is to be made, focused largely on the issue on the psychological validity of current proposals. As we shall see, CF research like many areas of XAI, claims that its intuition-based computational evaluations are proxies for psychological evaluations when such claims are at best, unsubstantiated, and, at worst, may be wrong.

## 1.1 Why Counterfactuals?

Intuitively, counterfactual explanations seem to convey a lot more information about a decision, prediction, or classification than factual explanations. To re-run the classic example, a loan refusal by an automated AI system using a *factual explanation* might explain the decision to a customer by saying "you did not get the loan because your profile is similar to applicants who were also refused" [Kenny and Keane, 2019; Kenny *et al*., 2021]. In contrast, a *counterfactual explanation* might explain the same decision by saying "if you earned $1k more, then you would have gotten the loan". Not only does this explanation inform end-users about key features leading to the decision, it also opens the door to "algorithmic recourse" allowing them to improve their chances of success; that is, the explanation may suggest "actionable" features that permit a remediation of the decision. Proponents of CFs have argued that these explanations accrue significant technical, psychological and legal benefits over other explanation techniques.

*Technically*, CF techniques build on a wave of popular XAI work showing that feature-importance analyses of ML models can provide automated explanations. Techniques – such as LIME and SHAP – showed that presenting information on the relative importance of features for a model's prediction can provide acceptable automated explanations [Ribeiro et al., 2016; Lundberg and Lee, 2017]. Some CF methods carries on directly from this approach in selecting features to change in the CF [Pedreschi *et al.,* 2019]. Wachter *et al.'s*, [2018] seminal work cast it as an optimisation problem balancing the proximity of CF to the test-instance against its proximity to the decision boundary. These works demonstrated the technical feasibility of finding automated explanations that appeared to be plausible to users.

*Psychologically*, this computational work was backed by a long-standing literature in Psychology [Byrne, 2007, 2019; Mueller *et al.,* 2019] and Philosophy [Woodward, 2005; Lewis, 2013] arguing for the centrality of counterfactuals in human cognition, explanation and science. But, it was Miller's [2019] seminal review of this literature that perhaps convinced the XAI community of the importance of these explanations. These works solidified the view that CFs could provide psychologically intuitive, plausible explanations; especially, if they were *sparse* (had few feature differences) and *proximate* (the *closest possible world*). However, as we shall see, solid evidence for many of these claims was not always forthcoming.

Legally, the final fillip to the counterfactual program came from the argument that CF explanations were compliant with the EU's General Data Protection Regulation (GDPR) [Wachter *et al.,* 2018]; so, they provided an explanation method that appeared to meet emerging regulatory requirements on AI systems. But, note, this claim rests on the assumption that people find CF explanations comprehensible, which remains to be proven sufficiently.

## 1.2 Motivation & Novel Contributions

In 2020, two reviews of the CF XAI literature, between them, surveyed 52 CF techniques [Karimi *et al.*, 2020a; Verma et al., 2020]. However, given the breakneck pace of this area, we have uncovered a further 48 CF-method-papers not referenced in these reviews (see Figure 1). Furthermore, these surveys do not focus on user studies, but rather on the technical features of explanatory methods. Thus, the present survey complements these earlier surveys, but offers a very different analysis of the area focusing on the key insights made, user studies on CF XAI, and the evaluation deficits in the field. As such the key contributions of the paper are:

- A novel, critical (updated) review of 100 CF XAI methods, focused on -- psychological (user studies) and computational (metrics) -- evaluation deficits
- The quantification of these deficits, to show the extent of underperformance in the research effort, to-date
- The proposal of a roadmap to standardize future evaluations and place them on a firmer psychological footing for scientific progress to be achieved.

As we shall see, the current critique rests on identifying the paucity of user studies on CF XAI (see Figure 1). This neglect undermines the "proxy" computational evaluations currently used, as they are intuition-based and not adequately ground-truthed. In the next section, we summarize the main "insights" driving CF XAI research (section 2). Then, with this analysis as a guide, we detail five main deficits in the area covering (i) the lack of user studies (section 3), (ii) the definition of plausibility (section 4), (iii) the issue of sparcity (section 5), (iv) the assessment of coverage (section 6) and (v) comparative testing (section 7). We conclude with proposals on how to rectify these deficits in future CF XAI work with a roadmap and benchmark evaluation metrics (section 8).

## 2 Counterfactual Insights

The previous reviews of CF XAI focus mainly profile techniques in terms of their technical properties. We offer a different analysis based on the high-level "insights" that have driven the area. We use this discovery-based approach as it forms the basis for our proposed standardized evaluation metrics; specifically, that key evaluation metrics should address the key "insights" that have advanced the field (see section 8). Although, there is huge diversity in the 100 distinct CF methods surveyed here, there are perhaps four big ideas that have driven the area forward: namely, that generated CF explanations need to be (i) guided by proximity, (ii) feature focused, (iii) distributionally faithful and, possibly, (iv) instance-based. Not all methods sign-up to these insights but many current proposals are designed to meet them as high-level requirements. They are summarized below.

*CFs are proximity-guided*. As we have seen, the seminal work on CF explanation [Wacheter *et al.*, 2018; Mittelstadt *et al.*, 2019] proposes perturbing the features of synthetic CF instances, under a loss function balancing proximity to the test-instance against proximity to the decision boundary for the CF class, using a scaled $L_1$-norm distance-metric. This idea has inspired follow-on work using different distance metrics (e.g., $L_2$-norm) or, indeed, combinations of distance metrics [Dandl *et al*., 2020; Artelt and Hammer, 2020], with added constraints to deliver *diverse* CFs [Mothilal *et al.,* 2020; Russell, 2019]. Hence, later, we will argue for specific distance metrics to benchmark evaluations (ideally, ones that are psychologically grounded).

*CFs need to focus on features*. The second major insight that quickly emerged in the area, was on the importance of focusing on the "right" features (actionable ones) to perturb and avoiding the "wrong" ones (immutable ones), by using *predictive importance* [Maartens and Provost, 2014; McGrath et al., 2018; Guidotti et al. 2018; Pedreschi *et al.,* 2019], "actionablility" [Ustun et al., 2019; Karimi *et al.*, 2020b], "coherence" [Russell, 2019; Gomez *et al.,* 2020] and "causality" [Karimi *et al.*, 2020c]; while also considering dependencies between features [Mothilal *et al.,* 2020; Kanamori *et al*., 2020]. In our evaluation guidelines we discuss some of the issues around evaluating this insight.

*CFs are distributionally-faithful.* It also became clear that the CFs relationship to the dataset distribution was also critical to plausibility, as some methods produced out-of-distribution (OOD) invalid data-points [Wachter et al, 2018; Laugel *et al.,* 2019] (see Figure 2). Hence, methods emerged using generative models [Dhurandhar *et al.,* 2018; Joshi *et al.,* 2019; Liu *et al.,* 2019; Singla *et al.,* 2019] and/or techniques manipulating latent features from deep learning models [Hendricks *et al.,* 2018; Van Looveren and Klaise, 2019; Akula *et al.,* 2020; Kenny and Keane, 2021]. So, benchmark metrics need to assess the distributional properties of CFs and their coverage.

*CFs are instance-guided:* Finally, some have argued that the best way to generate "good" CFs is to rely on the dataset, either (a) directly by using Nearest Unlike Neighbours (NUNs) [Nugent et al., 2009] or (b) indirectly, by adapting instances [Goyal *et al.,* 2019; Keane and Smyth, 2020; Delaney *et al.,* 2020; Smyth and Keane, 2021]; closely related methods preferentially select synthetic CFs falling in dense regions of the dataset [Poyiadzi *et al*., 2020]. We propose a relative-distance metric to capture this.

In the following five sections, we critique the psychological and computational evaluations used in CF XAI, before considering how the deficits found might be rectified.

## 3 Deficit #1: Neglecting Users

The neglect of user studies is the "original sin" of XAI research. In an XAI-wide survey, Adadi and Berrada [2018] report that only 5% of papers evaluated interpretability. In the CF XAI papers surveyed here, only 31% of papers perform user studies (36 out of 117) and fewer (21%) directly user-test a specific CF method (i.e., many are non-model tests). Indeed, as the area expands exponentially the relative proportion of user studies is decreasing (see Figure 1). Furthermore, many of these studies are methodologically questionable (e.g., use low Ns, poor or inappropriate statistics, unreproducible designs). Many user studies test the use of CFs as explanations relative to no-explanation controls, rather than testing the specific methods. So, to use a blunt metaphor, we may be fine-tuning AI methods with elaborate bells and whistles that no human-ear can hear.

To be more positive, most studies reporting these tests do show CFs to be useful and sometimes preferred by end users (e.g., [Lim *et al.,* 2009; Dodge *et al.,* 2019]). Lim *et al.* [2009] tested the use of What-if, Why-Not, How-to and Why explanations and found that they all improved performance relative to no-explanation controls. Dodge *et al.* [2019] assessed four different explanation strategies (e.g., case based, counterfactual, factual) on biased/unbiased classifiers and found counterfactual explanations to be the most impactful. However, it should also be said, that some studies show that CFs explanations often require greater cognitive effort and do not always outperform other methods [Lim *et al.,* 2009; Lage *et al.,* 2019; van der Waa *et al.,* 2021].

Notably, however, few of these studies directly test a particular CF method. We have found that only 25 papers (out of 100 method papers) perform user studies on the proposed CF method (e.g., [Goyal *et al*., 2019; Singla *et al*., 2019; Lucic *et al.,* 2020]). These studies typically show some improvement in people's performance on or judgement of an AI system, relative to no-explanation controls. Finally, even fewer of these studies pit one method against another in a comparative user study (only 7 out of 100 method papers; e.g., [Akula, *et al.,* 2020; Förster *et al.,* 2020a, 2020b]).

In summary, this means that from 100 distinct CF algorithms in the literature, only 7% report user studies that specifically address the detailed properties of their methods. This state-of-affairs has knock-on effects for other aspects of CF evaluation. Basically, we do not know whether one method is better than another on (i) plausibility (section 4), (ii) sparsity (section 5), or (iii) coverage of representative problems (section 6). In the following sections, we detail these deficits in the evaluation of CF techniques.

## 4 Deficit #2: What's Plausible?

All counterfactual explanation methods make the core claim that their method delivers *plausible* or *relevant* or *feasible* or *helpful* or *good* explanations to human users, explanations that meet user/stakeholder goals in diverse AI systems. Different techniques define plausibility in different ways, depending on the insights they promote (see section 3): so "plausible" CFs (i) are close the test instances and the decision boundary, (ii) use the "right" features (e.g., mutable, actional, causal) or (iii) are faithful to the distribution, (iv) arise from the training data. For the most part, these claims are not supported by user tests but are declared on intuitive grounds. Support for these claims is provided by "proxy" computational evaluations that often directly parallel the theoretical claim (e.g., a generative technique is plausible because it produces within-distribution CFs)[2]. However, all of these claims need to be grounded psychologically. Here, we briefly consider what this means for the plausibility-as-proximity and plausibility-as-good-features claims.

*Plausibility-As-Proximity*[3]. Many techniques argue that the instances in the CF need to be close [Wachter *et al.*, 2018]. But, aligning similarity with plausibility is problematic. First, there is no concrete evidence – as in human ratings studies -- showing that plausible CFs are more similar. Second, reporting that the counterfactuals generated involve highly-similar pairs invites the question: "how close do they have to be, to be plausible?". A CF with a low distance-score could well be incomprehensible if it violates common-sense (e.g., a house with 2.312 rooms might be very close to the target house of 2 rooms but meaningless; see also Figure 2). Third, in computational evaluations a wide range of different distance metrics are used, which raises the question as to which one is the right one (is it $L_1$, $L_2$ or some other variant?). So, though papers report distance metrics for computational evaluations because these metrics are not grounded psychologically, they do not tell us whether people will find them acceptable. We know of no user studies that confirm which distance metric is a good proxy for the psychological distance of CF explanations (e.g., something akin to Wang et al.'s, [2004] structural similarity index (SSIM) for images). If similarity is to be used as a proxy for plausibility, the chosen metric needs to be psychologically grounded (and explicitly, linked to people's assessments of plausibility).

*Plausibility As More-Good-Features*. Others argue for the importance of "good" features for plausibility. But, aligning the plausibility of CFs with the appropriate use of "more good features" is also problematic. First, as with similarity, we have no direct evidence on the link between plausibility and the feature-characteristics of CFs. Though, intuitively, we may agree that immutable features should not occur in plausible CFs, we do not know how different features-types – actionable, mutable, feasible or causal -- impact people's perception of explanations; for example, how many "good" features are required for a plausible CF or how does the acceptability of explanations change with variations in a user's causal model of the domain and so on. Note, that [Lakkaraju and Bastani , 2020] recently found that people can be *misled* into trusting a model more if the factual explanations used avoid "sensitive" features (such as gender and race). So, without thorough user studies, the plausibility of CFs with more-good-features remains an open question.

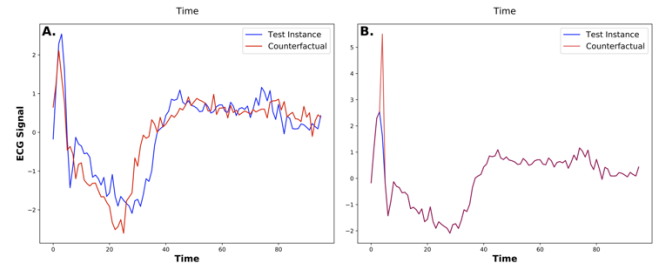

Figure 2: Two counterfactual comparisons for a time-series EEG signal (A) a "native CF" (a NUN from the dataset) that is distant from the test instance but within distribution ($L_1$-norm = 38.29) and (B) a synthetic CF generated by a proximity method [Wachter et al., 2018] which is much closer to the test instance ($L_1$-norm = 3.88) and offers a sparser solution but is out-of-distribution according to the OCSVM metric (given the spike). Neither of these counterfactuals is likely to be plausible to end-users (from [Delaney et al., 2020]).

## 5 Deficit #3: The Shape of Sparsity

It is generally agreed that "good" CF explanations will be *sparse* (aka have few feature differences). This sparsity claim is made on intuition, sometimes with a hand-wave to psychological research on working memory limitations, visual memory, or limits on human category learning. An automated explanation for a house price prediction saying "if it had two more rooms and a bigger garden, it would be double the price" is intuitively viewed as being better than one saying "if it had 2 more rooms, a bigger garden, a larger bathroom, a wider entrance-hall and a two-car garage, it would be double the price". But, how sparse should sparse be? Most researchers do not commit to an optimal sparsity level for "good" CFs. The few papers that do commit, vary widely in their positions, from 1-2 feature-differences [Schleich et al., 2021; Keane and Smyth, 2020] to 8-30 feature-differences [McGrath et al., 2018; Martens and Provost, 2014]. Choosing such a level seems to be important as it looks like many popular methods do not find low-sparcity CFs [Schleich et al., 2021]. However, until recently, there was little hard evidence on what sparsity level might be psychologically optimal.

---

[2] For now, we will set aside the circularity of this sort of evaluation, but note that it is an issue to be considered.

[3] We also set aside the fact that the concept of *plausibility* is not well understood in Psychology; though, there is agreement that it tends to depend heavily on user's knowledge of a domain, rather than on similarity *per se* [Connell and Keane, 2007].

All the user studies that show some the general effect of CFs in XAI [Dodge et al., 2019; Lim et al., 2020] use single-feature-difference CFs, so they are silent on the impact of >1 feature-differences on people's responses. The first direct tests of sparcity come from a single group in the University of Ulm [Förster et al., 2020a, 2020b]. Over a series of studies, assessing CF methods, they systematically varied the median feature-differences in CF explanations shown to users. Interestingly, they found that people did not prefer 1-difference CFs and rated CF explanations with 2-3 feature-differences as being much better.

The point of this discussion is not all about what the optimal psychological sparcity-level might be, but is rather again to underscore the importance of using evaluation metrics that are backed by user studies. Clearly, the precise optimal sparcity level for people is likely to vary (e.g., in tabular versus text and image data, especially when chunking is taken into account)[4]; however, it is probably wise to settle on a low-end score, if only to set a high bar for distinguishing between different methods.

## 6   Deficit #4: Covering Coverage

Any adequate counterfactual method needs to guarantee that, on the whole, it will produce "good" counterfactual explanations and avoid "bad" ones over some set of representative problems. If a method generates implausible explanations, even some of the time, given people's tendency towards "algorithmic aversion" it will fail [Burton et al., 2019]. Laugel et al. [2019] showed that one type of "bad" CF (i.e., out-of-distribution items) can be as high as 36% for some CF methods and Delaney et al. [2020] have shown that even close, low-sparcity CFs can be out-of-distribution (see Figure 2). In the 100 systems reviewed here, we found that only 22% report "coverage results", though the definitions of the concept differ (e.g., see [Keane and Smyth, 2020; Schleich et al., 2021; Dandl et al, 2020]).

What we are calling *coverage* can be measured in a number of different ways. Some have used out-of-distribution (OOD) measures to track the numbers of invalid CFs being produced by models: including, using IM1 and IM2 [Van Looveren and Klaise, 2019], Local Outlier Factor (LOF) [Breunig et al., 2000] and One-Class-SVM (OCSVM) [Schölkopf et al., 1999]. Obviously, generative CF methods place a strong emphasis on staying within the distribution and, as such, have championed OOD evaluations [Joshi et al., 2019; Liu et al., 2019; Singla et al., 2019]. However, these measures tell us more about "bad" CF explanations than "good" ones. If we assume OOD-CFs are implausible (and this needs to be verified psychologically) then the remaining in-distribution CFs only "may" be plausible; being in-distribution is not in itself a guarantee of plausibility (see Figure 2). We need a measure of which of those in-distribution CFs that *are* plausible.

[Keane and Smyth, 2020] have proposed the idea of *explanatory competence* (by analogy to *predictive competence* [Juarez et al., 2018]) using a definition of *explanatory coverage* (*XP_Coverage*). Assume we have a function that captures psychologically-acceptable CF explanations -- *explains(c, c')* -- where, *c*, is the test instance and *c'* the CF-instance. Then, the *explanatory competence* of a dataset, *C,* can be represented by a *coverage set* (Eq. 1) and degree of explanatory competence is the size of the coverage set as a fraction of the dataset (Eq. 2):

$$XP\_Coverage\_Set(C) = \{c' \in C \mid \exists c \in C\text{-}\{c'\} \ \& \ explains(c, c')\} \quad (1)$$

$$XP\_Coverage(C) = |XP\_Coverage\_Set(C)| / |C| \quad (2)$$

This evaluation metric is critical as it gives us a metric for the likelihood of people encountering a poor explanation for a given method. However, an issue for this measure is how to define the *explains* function. Keane and Smyth [2020] adopted the simple expedient of defining it as any CF with ≤2 feature-differences; but, more complex, psychologically-backed definitions need to be adopted. As we shall see later, this type of measure needs to be part of any standard, benchmark evaluation.

## 7   Deficit #5: Comparative Testing

The final deficit to be noted about the CF literature is the lack of comparative testing in CF papers. We found only 40% (40 of 100 method papers) reported any form of comparative testing[5]. While the community has clearly moved beyond John McCarthy's "Look Ma, no hands !" phase, given the sheer number of different methods now in the literature, more comparative testing is clearly needed. Fortunately, more recent papers have a greater tendency to report such tests and many groups are making their code publicly available. We believe that with an agreed set of evaluation measures that are grounded in user tests, this field is poised to deliver solid scientific advances.

## 8   Roadmapping & Benchmarking

The human aspect of XAI research places a whole new set of constraints on the AI methods being developed to ensure fairness, accountability and trustworthiness. In this survey, we reviewed the evaluative shortcomings of XAI research on counterfactual explanations, mainly with an eye to providing a better psychological grounding for the area. To cement real progress we recommend a roadmap for future work and set of evaluative benchmarks to be adopted.

---

[4] Note, what counts as a feature will not always be an input-feature (as in tabular data); we may be dealing with latent/"semantic" features of the model, which may be *superpixels* for image data or *motifs* in time-series, and so on.

[5] Note, this number overestimates comparative tests as it includes papers that solely test variants of their own algorithm.

## 8.1 Roadmap for Psychological Grounding

We have argued that the core deficit facing CF XAI is the gap in user testing and the proper psychological grounding of computational evaluations. As such, our research roadmap recommends a program of general and specific testing of counterfactual explanations.

Broad user-testing needs to be carried across a range of diverse domains (e.g., decision making systems using datasets in different enterprise domains). For these general tests, we recommend the studies by [Dodge *et al.*, 2019] and [Lage *et al.*, 2020] as excellent experimental designs.

Specific user testing is also required to backstop the benchmark computational metrics to be used for comparative evaluations of methods. These studies will need to determine (i) the relationship between proximity and "good" CF explanations as determined by human-users and the most suitable distance metric to approximate this (whether it be $L_1$, $L_2$, or some other metric), (ii) the bounds on sparcity (to determine the optimal frequencies in feature-differences for comprehensibility, in different domains), (iii) how different classes of feature-differences are cognitively appraised (e.g., mutable/non-mutable, causal, actionable) and how this interacts with expertise in a domain, (iv) whether people can spot out-of-distribution CFs, how they appraise them and the cognitive factors affect these appraisals. For these specific tests, we recommend [Förster, et al., 2020a, 2020b] and [van der Waa et al, 2021] for the best experimental designs.

## 8.2 Benchmaking Evaluative Methods

As part of this roadmap we also need standardized set of "proxy" computational evaluation metrics to help us decide between the 100-odd methods in the literature. We propose four benchmarking metrics that are selected to address the key insights guiding CF XAI research: including, benchmark metrics for proximity, sparcity, coverage and relative distance. We do not list benchmark datasets as there is already good agreement across papers on the key ones to use.

*Proximity.* Distance metrics are a rough, but reasonable, proxy for the overall performance of a CF method, though they can hide a lot; by convention in the literature has commonly used $L_1$- and $L_2$-norms. To enable retrospective comparisons the $L_1$-norm probably has to be used, alongside the $L_2$-norm, until it is clear which is the more psychologically-valid measure. Karimi et al. [2020b] provide a good example of this sort of evaluation.

*Sparcity.* We have seen that sparcity is a critical index of the psychological acceptability of CF explanations. As distance metrics report averages, tests need to be broken out by sparcity levels showing the frequency of CFs generated; the range from 1-5 feature-differences seems critical, as it may well discriminate models (e.g., a model producing most of its CFs with >4 differences may be questionable). Schleich *et al.* [2021] give a good example of such reporting.

*Coverage.* Armed with a definition of "good counterfactuals" (even a rough one any CF with ≤3 feature-differences) a coverage metric capturing the proportion of good CFs for test-sets will provide a global estimate of the adequacy of methods; showing how likely they are to generate unacceptable CFs. Smyth and Keane [2021] provide a good example of this sort of reporting. Allied to this, we also need a measure of "bad" CFs being generated, using an OOD measure; we recommend the Local Outlier Factor (LOF) metric, for now, used in some papers [Breunig et al., 2000].

*Relative Distance.* Finally, we propose a relative-distance measure (with the most psychologically-valid metric) comparing the mean distance of CF-pairs (between the test and CF instance) over the mean distance of "native counterfactuals" (NUNs). This metric is included to assess the instance-guided insight, as it shows whether the CFs generated by a method are closer than "natural" CFs in the dataset, on the assumption (to be tested) that this makes them better (see [Smyth and Keane, 2021] for an example).

## 8.3 Caveats & Conclusions

We should conclude with three caverats about these proposals that are important. First, we recognize that there are wider issues about the use and assessment of CF algorithms that are not considered here; namely, issues around the fidelity of explanations to a model's predictions, the robustness of methods, the speed/complexity of the algorithm, ethical and stakeholder issues and so on (see [Sokol and Flach, 2020], for a long list of requirements). Here, we have focused on, what we see as, core issues about the psychological and computational evaluation of these methods. All of the other requirements are also important.

Second, we are conscious that we have *not* proposed a benchmark metric for featural aspects of CF explanations; for example, whether the CFs generated by a method observe mutability, actionability and causal constraints. This aspect of CF methods also needs to be evaluated in a standardized way but there is less agreement in the literature on how this might be done. Furthermore, any broadly-applicable metric will require a dataset-by-dataset agreement on lists of immutable/mutable features, actionable features and also, possibly, causal models. While it is not impossible to do this, it does require greater commitment to agreement in community. For potential candidates for such evaluations see [Ustun *et al.*, 2019; Karimi et al., 2020b; Schleich *et al.*, 2021].

Third and finally, we do not underestimate how difficult it will be to rectify the deficits identified here and how much work will be required by inter-disciplinary teams to address these issues. However, if XAI does not address these sorts of issues then the emerging impediments to the widespread deployment of AI systems will, simply, not be overcome. And, perhaps, more fundamentally, we will not see sufficient scientific progress in these parts of the AI firmament.

## Acknowledgements

This publication has emanated from research conducted with the financial support of (i) Science Foundation Ireland (SFI) to the *Insight Centre for Data Analytics* under Grant Number 12/RC/2289_P2 and (ii) SFI and the Department of Agriculture, Food and Marine on behalf of the Government of Ireland to the *VistaMilk SFI Research Centre* under Grant Number 16/RC/3835.

# Annotated Bibliography

*     means cited in paper
U     means user study not cited but reviewed
M     means model not cited but reviewed
NonCF     means not about CFs
REV     means review/survey article
EVAL     means article on evaluation metric
No Prefix means not cited in paper, but surveyed